\documentclass[letterpaper, 10 pt, conference]{ieeeconf}

\IEEEoverridecommandlockouts               %

\usepackage[style=ieee,mincitenames=1,maxcitenames=2,maxbibnames=12,doi=false]{biblatex}
\renewcommand*{\bibfont}{\footnotesize}

\usepackage{amsmath}
\usepackage{graphicx}
\usepackage{subcaption}
\usepackage{url}
\usepackage{csquotes}
\usepackage[american]{babel}
\usepackage{siunitx}

\usepackage{tikz}
\usepackage{hyperref}
\newcommand\copyrighttext{%
  \footnotesize \textcopyright 2023 IEEE. Personal use of this material is permitted.
  Permission from IEEE must be obtained for all other uses, in any current or future
  media, including reprinting/republishing this material for advertising or promotional
  purposes, creating new collective works, for resale or redistribution to servers or
  lists, or reuse of any copyrighted component of this work in other works.
  DOI: Accepted at ICRA 2023 %
  }
\newcommand\copyrightnotice{%
\begin{tikzpicture}[remember picture,overlay]
\node[anchor=south,yshift=10pt] at (current page.south) {\fbox{\parbox{\dimexpr\textwidth-\fboxsep-\fboxrule\relax}{\copyrighttext}}};
\end{tikzpicture}%
}

\title{\LARGE \bf Buoyancy enabled autonomous underwater construction with cement blocks}
\author{
\authorblockN{Samuel Lensgraf\authorrefmark{1}, Devin Balkcom\authorrefmark{1}, Alberto Quattrini Li\authorrefmark{1}}
\authorblockA{Dartmouth College\authorrefmark{1}}
\thanks{We would like to thank Amy Sniffen and Julien Blanchet for experimental support, and Eren Aldemir and Amanda Sun for the support in the robot hardware development. This project was partially supported by the NSF GRFP, CNS-1919647, 2024541, 2144624.}
}

\begin{document}
\maketitle
\copyrightnotice
\begin{abstract}
    We present the first free-floating autonomous underwater construction system capable of using active ballasting to transport cement building blocks efficiently. It is the first free-floating autonomous construction robot to use a paired set of resources: compressed air for buoyancy and a battery for thrusters. In construction trials, our system built structures of up to 12 components and weighing up to \SI{100}{Kg} (\SI{75}{Kg} in water). Our system achieves this performance by combining a novel one-degree-of-freedom manipulator, a novel two-component cement block construction system that corrects errors in placement, and a simple active ballasting system combined with compliant placement and grasp behaviors. The passive error correcting components of the system minimize the required complexity in sensing and control. We also explore the problem of buoyancy allocation for building structures at scale by defining a convex program which allocates buoyancy to minimize the predicted energy cost for transporting blocks.
\end{abstract}

\section{Introduction}

Near coast underwater infrastructure plays an important role in many of the most basic aspects of society. The United Nations estimates that about half of the world's seafood comes from aquaculture~\cite{NOAA2020Fisheries}. Offshore wind energy currently produces \SI{42}{MW} of electricity in the U.S. alone with numerous projects expected to expand that capacity~\cite{DOE2022OffshoreWind}. While autonomous underwater vehicles (AUVs) have been widely explored for aiding with inspection and exploration tasks~\cite{petillot2019underwater}, little work has been done to explore using them to directly aid in constructing underwater infrastructure.

\begin{figure}[t]
    \centering
    \includegraphics[width=0.85\linewidth]{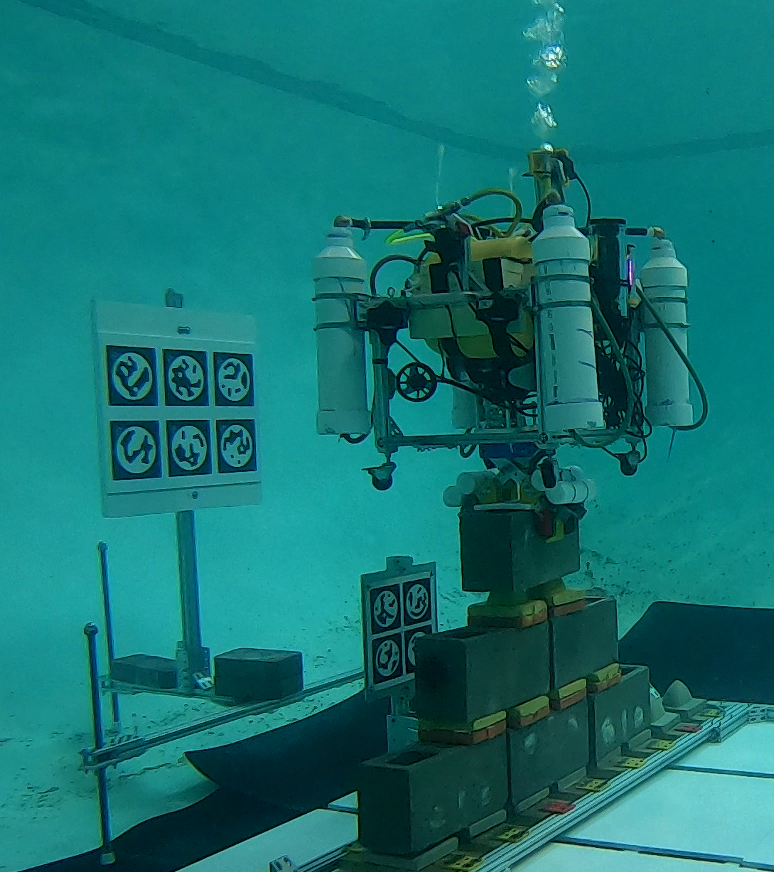}
    \caption{Placing the final block on a pyramid while releasing excess buoyancy.}
    \label{fig:hero-image}
\end{figure}

Autonomous construction in water presents the unique opportunity of controlling the construction vehicle's buoyancy, which allows an AUV to build heavier and larger structures on limited battery capacity than drone-based systems. To exploit this opportunity, we developed the first free-floating autonomous construction system that actively tunes its buoyancy, allowing it to manipulate cement building blocks efficiently. Figure~\ref{fig:hero-image} shows our system placing a cement block on top of a 2D pyramid. %

Our AUV system is wholly designed around the task of constructing cement block structures. It consists of several novel components: a novel one degree-of-freedom manipulator that allows simple grasp behaviors which align the AUV, a novel two-component cement building block system designed specifically to accept large amounts of placement error, and a simple active ballasting system and associated control behaviors that offset variable amounts of weight.

While it is common to design long term autonomous systems with backup energy sources, the balancing of two complementary and distinct resources during a manipulation task is, to our knowledge, unexplored. We defined a convex program which captures the trade off between battery power and compressed air. The convex program can be used to plan buoyancy allocations for large structures. We use this convex programming formulation to explore the problem of scale more deeply than possible in our physical experiments. 

To alleviate positioning and localization errors, our system builds structures with slightly modified cement blocks combined with molded cement interlocking elements, referred to as \textit{cones}. Structures are built of alternating layers of cement blocks and cones which ensure that each layer helps the next slide into place. Our AUV's novel manipulator is fitted with two phases of error correction which when combined with our novel compliant grasping strategies allows the AUV to slide into the proper alignment as it grasps a component.

This work represents a first step towards large scale construction. A number of future challenges, including robust perception and adaptation to external disturbances, will be study of future work, as discussed at the end of the paper.
\section{Related Work}

In our previous work~\cite{Lensgraf2021DropletI}, we developed the basis used for controlling and localizing our AUV. The previous iteration of our system created a structure of eight smooth, uniformly shaped, nearly neutrally buoyant blocks. Our current system built a structure of twelve components of two different shapes that are significantly negatively buoyant and high friction. 

We presented preliminary work on our system at the ICRA 2022 construction robotics workshop~\cite{Lensgraf2022StoneClawWorkshop}. We presented preliminary results using the buoyancy allocation convex program but did not apply it to the question of scale. The system contained basic versions of several of the components described here, but the control software and hardware were not capable of stacking blocks.

Other autonomous underwater construction systems have primarily focused on tele-operation such as the haptic feedback system for operating a back hoe developed by \textcite{hirabayashi2006teleoperation}. Augmented reality has been explored as a way to manipulate large objects underwater using waterproofed construction equipment \cite{Utsumi2002DevelopmentFT}. Surface-level self propelled blocks have also been explored for building bridge-like structures \cite{seo2016assembly}. One of our goals in designing the block the robot builds with was to keep them simple and similar to construction materials that are easily available. Self assembly of robotic systems in water has also been explored~\cite{ganesan2016stochastic,vasilescu2005autonomous,tolley2011programmable}.

\textbf{Land-based construction systems.} Land-based robotic construction systems have seen more development than air and water-based systems~\cite{saidi-construction}. Land-based systems have used a variety of mobility designs including wheeled robots~\cite{saboia2019autonomous,gambao1997robot,hamner2010autonomous} and tracked robots~\cite{dorfler2016mobile,helm2012mobile}. Robotic systems for autonomously laying brick walls have been explored since the birth of autonomous construction research~\cite{andres1994Rocco,balaguer-site}. Mobile-base 3D printing robots are currently being explored both in industry and in research~\cite{sustarevas2018map, ApisCorWebSite}. Land-based robotic construction systems typically assume easy access to a power source or spare batteries, limiting the need for explicitly considering energy use during construction.

\textbf{Drone-based construction systems.} Latteur \textit{et al.} explored using drones to stack interlocking cement blocks~\cite{Latteur2016MasonryCW, Latteur2018MasonryFeasibility}. While the problems of designing easy to assemble cement blocks and localizing the drone were explored, this work was tested using human pilots. The problem of battery capacity's effect on scale was left as future work. 

The largest structure assembled by drones appears in the Flight Assembled Architecture Installation \cite{augugliaro2014flight} in which a team of drones manipulated 90 gram foam blocks. Because the construction process centers around using a large team of UAVs, the UAVs can be easily swapped out. This eliminated the need for explicitly considering energy usage. The construction of truss structures using teams of quadrotor drones was explored by \textcite{lindsey2012construction} and in simulation by \textcite{dos2018iterative}. Using drones as the base of an aerial 3D printing system has also been explored~\cite{hunt20143d, Dams202Drone3DPrint}.

\textbf{Autonomous underwater manipulation.} Autonomous underwater manipulation systems, referred to as intervention AUVs, are often designed to be general purpose agents fitted with complex, high degree-of-freedom manipulators for performing tasks such as manipulating a panel or collecting samples~\cite{capocci2017inspection, Cieslak2015PanelManip}. Problems such as station keeping while manipulating an object using a high-degree-of-freedom manipulator require complex control strategies~\cite{sanz2012trident, Simetti2014FloatingManip}. Manipulating objects using teams of AUVs has also been explored~\cite{simetti2017manipulation}. Most underwater manipulators are prohibitively expensive. Even simple models with more than one degree of freedom can cost tens of thousands of US dollars~\cite{sivvcev2018underwater} and range up to millions of US dollars.

To overcome the cost and system complexity associated with most general purpose underwater manipulators, we designed our system to be as simple as possible while still achieving the task at hand. The most closely related autonomous underwater manipulation system to ours is the system by \textcite{palomeras2014autonomous} in which the AUV docks in a specially designed mount before executing a manipulation task by forcing rods into cones mounted above the panel. This work in part inspired the compliant plunging procedures used to grasp objects with our AUV.

\textbf{Buoyant robots.} Active ballasting has long been used by autonomous and remotely operated underwater vehicles. Compressed air coming from an above-water source was used to offset the weight of a payload during a recovery procedure~\cite{Wasserman2003MATERov}. Other AUVs have exploited compressed air based active ballast to accommodate changes in salinity and thereby buoyancy in estuary environments~\cite{Love2003BuoyancyControl}. 

Piston tank and pump based active ballasting systems are commonly used in underwater gliders and submarines~\cite{Zhang2014UWGlider, Bokser2004MiniSub}. \textcite{detweiler2009saving} designed a robotic platform for repeatedly accommodating dynamic payloads using a piston-tank-based active ballasting system and compensated for changing centers of mass by moving the robot's battery. Their system could accommodate up to \SI{1}{kg} of payload. They explored the trade off between using buoyancy and thrusters, however the model is not used for planning.

\section{AUV construction system}

Our AUV construction system is a low-cost AUV designed specifically for construction with cement blocks. Its hardware and software are co-designed to achieve robust assembly while keeping complexity low.

\subsection{Error correcting cement blocks}

\begin{figure}[h]
    \centering
    \includegraphics[width=0.95 \linewidth]{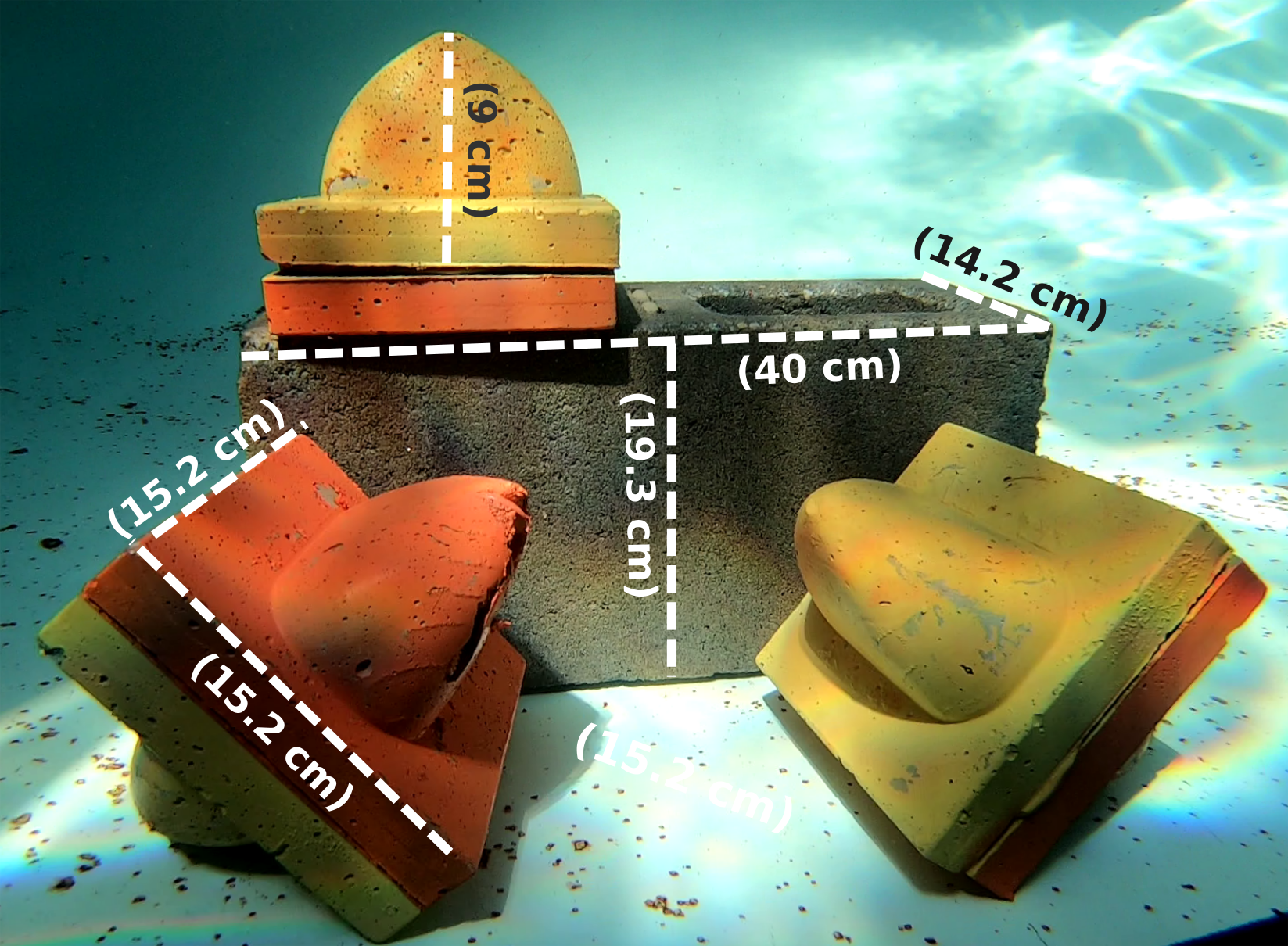}
    \caption{Interlocking cone inserts (yellow and orange) and rectangular cement block.}
    \label{fig:cones_and_blocks}
\end{figure}

Our AUV builds structures using a novel two-component process in which layers of error correcting cone inserts provide passive error correction when inserted between layers of standard, commercially available rectangular cement blocks. To facilitate sliding error correction, we ground slight chamfers into the sides of the internal holes of the cement blocks. The cone inserts weigh \SI{3.9}{Kg} (\SI{3.2}{Kg} in water) and the rectangular blocks weigh \SI{12.9}{kg} (\SI{9.5}{Kg} in water). Figure~\ref{fig:cones_and_blocks} shows the cones and blocks.

The cone inserts are made of two part molded cement. The top half (yellow in Figure~\ref{fig:cones_and_blocks}) is made of a 30\% by volume perlite mix and the bottom half (orange in Figure~\ref{fig:cones_and_blocks}) is embedded with bolts in the tip. This creates an asymmetric weighting that helps the cones tend to fall in the proper orientation through water. The base of the cones are slightly wider than the cement blocks, which allows them to easily be grasped when resting in the cement blocks.

Based on the CAD design, the cones can correct for up to \SI{5}{cm} of position error on the y-axis and \SI{2.5}{cm} of error on the x-axis where the y-axis is along the long dimension of the block and the x-axis is along the shorter side looking down.

\subsection{Manipulator}

\begin{figure}[h]
    \centering
    \includegraphics[width=0.95 \linewidth]{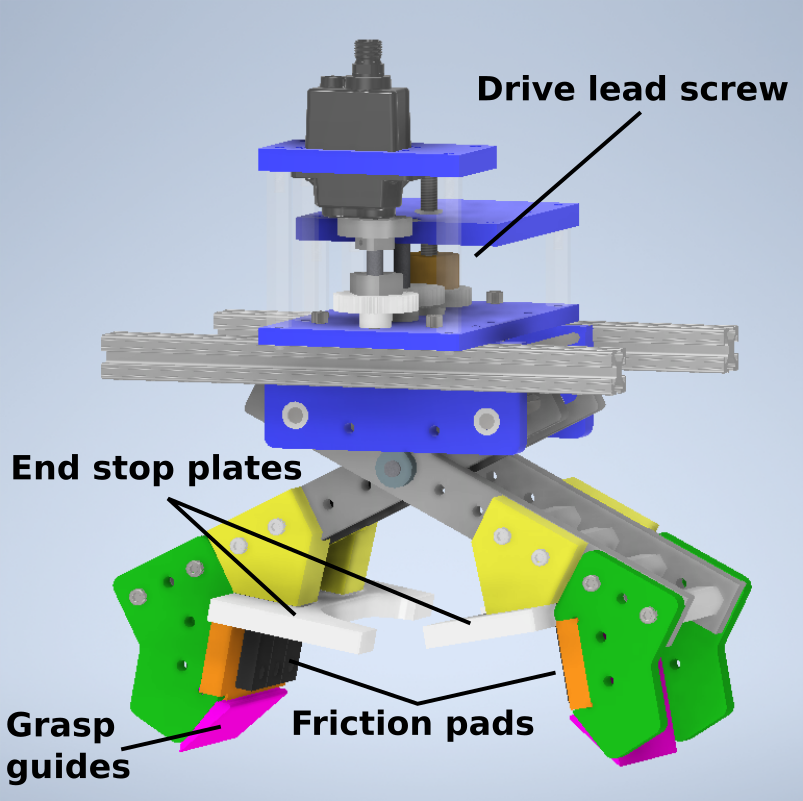}
    \caption{CAD rendering of error correcting 1DOF manipulator.}
    \label{fig:manipulator_rendering}
\end{figure}

The construction AUV is fitted with a purpose designed one degree-of-freedom (DOF) manipulator. The primary linkage of the manipulator is inspired by centuries-old stone grabber designs in which a pair of jaws exploit the weight of the item they grasp to draw the claw more strongly closed.

To drive the jaws of the manipulator, a high power, depth rated servo\footnote{\url{https://www.bluetrailengineering.com/servos}} drives a lead screw nut which is forced against thrust bearings between two spaced plates (blue in Figure \ref{fig:manipulator_rendering}). Enough space is left between the two plates that the nut can travel without spinning, allowing gravity to draw the jaws closed. When the manipulator reaches the end of its motion, a relay is switched off to prevent the stalled servo from drawing additional current.

When opened, the shape of the manipulator forms a triangle with two end stop plates (horizontal white plates in Figure \ref{fig:manipulator_rendering}). The \SI{26}{cm} wide triangular opening can be forced down on an object, pushing the end stop plates against its flat upper surface to correct error along the z-axis. The triangle opens wide enough that about \SI{6}{cm} of error along the block's x-axis can be tolerated. 

The end stop plates are shaped to fit around the error correcting cones, allowing improved error correction. The end stop plates can accept up to \SI{6}{cm} of error on the x-axis and \SI{7}{cm} of error on the y-axis and still slide on the proper surface of the cone. See Figure~\ref{fig:sliding-grasps} for an example of how the end stop plates are used. The manipulator is not capable of correcting grasp error for the cement blocks along their y-axis. This problem is mitigated by the high error tolerance of the cones on that axis.

\subsection{Active Ballasting}
\label{subsec:active-ballasting}

\begin{figure}[h!]
    \centering
    \includegraphics[width=0.75 \linewidth]{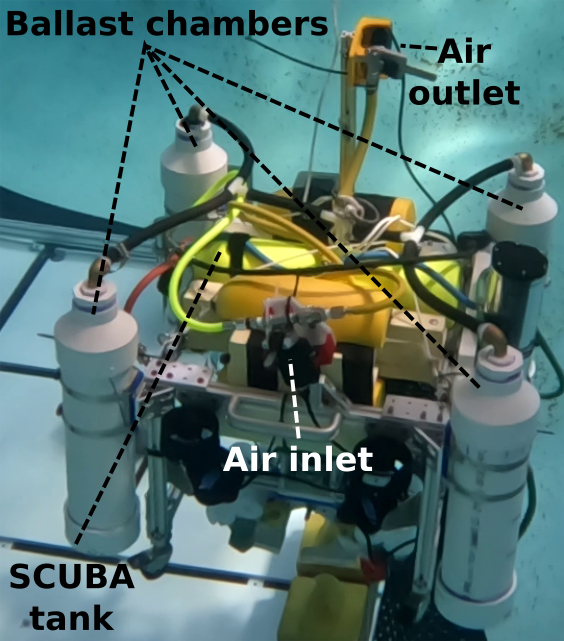}
    \caption{Active ballasting system.}
    \label{fig:active_ballast_system}
\end{figure}

To allow our AUV to manipulate the heavy cement blocks without stressing its electronics,
we designed a simple active ballasting system that offsets large amounts of weight using compressed air. To lift a single cement block weighing \SI{9.5}{kg} in water with thrusters alone, the vehicle must pull about 470 watts from its 230 watt-hour battery. At this rate, the vehicle would be able to operate for about thirty minutes. This would be impractical for building structures of any size. 

Servo actuated inlet and outlet valves allow the AUV to release compressed air from a 3 cubic liter SCUBA tank pressurized at \SI{3000}{PSI}. The SCUBA tank stores enough compressed air to offset roughly \SI{600}{kg} of water at 5 meters deep. Four ballast chambers made of four inch PVC pipe store air at the ambient pressure to increase the vehicle's buoyancy. The ballast chambers are oriented vertically to limit sloshing when partially filled and are spaced far apart to limit the pendulum effect of carrying a large mass below the AUV. Figure \ref{fig:active_ballast_system} shows the components of the system.

To change its buoyancy, the AUV translates a scalar tank level, $b \in [0.0,1.0]$, into a fixed downward thrust. $b = 0.0$ corresponds to enough thrust to lift the block without buoyancy and $b = 1.0$ corresponds to nearly no assistance from the thrusters.

As the pressure in the SCUBA tank decreases, so does the flow rate of air into the ballast chambers. To give $b$ a uniform meaning as pressure decreases, we use a pulsing strategy to release air into the ballast chambers. After grasping a block, the air inlet valve is pulsed until the vehicle ascends with the combined force of positive buoyancy and thrusters.

\subsection{Construction process}

\begin{figure}[h]
    \centering
    \includegraphics[width=0.9 \linewidth]{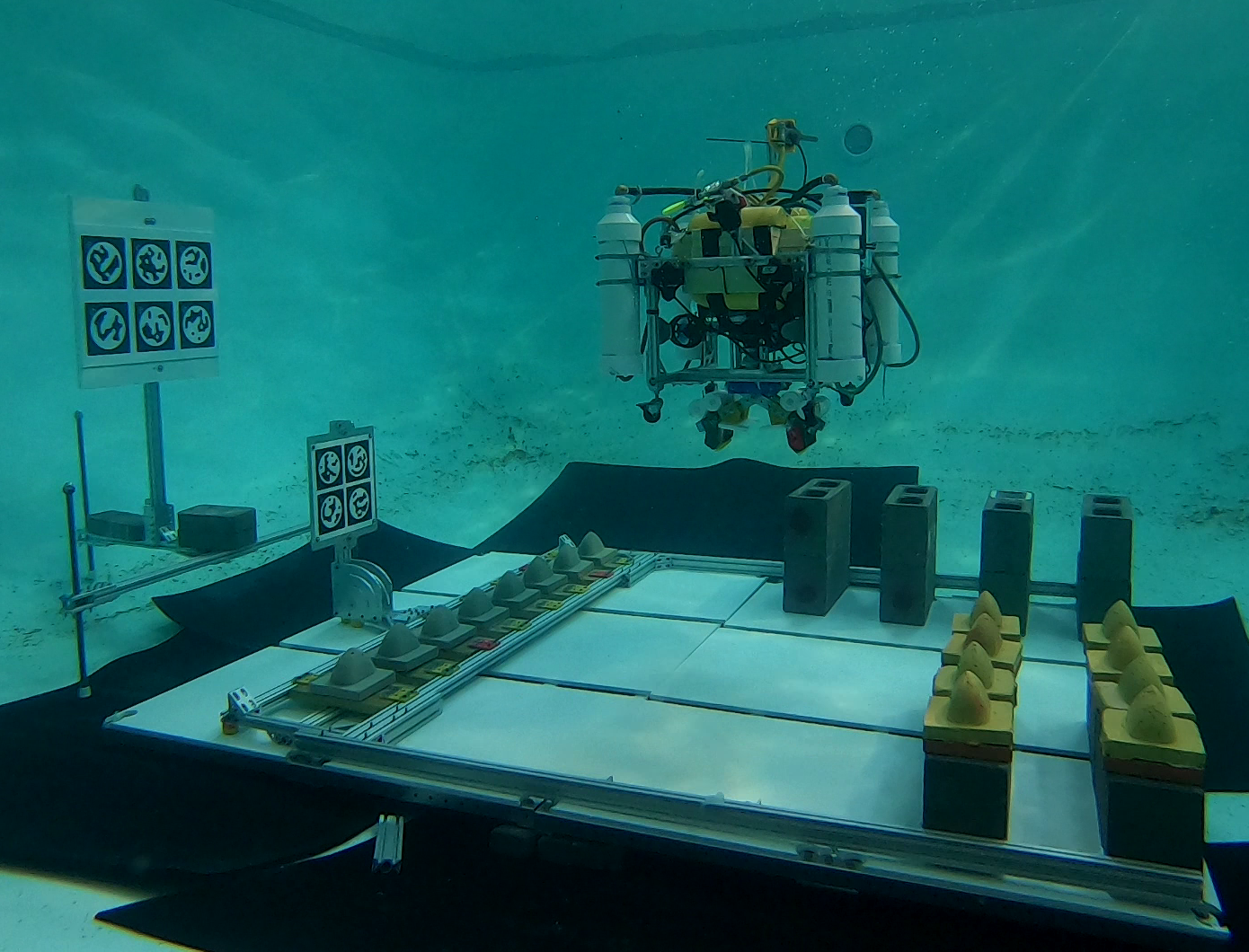}
    \caption{Blocks and cones arrayed in a pre-construction configuration.}
    \label{fig:construction_area_wide_view}
\end{figure}

The construction process is encoded as a set of behaviors paired with way points specified in a global coordinate frame. Blocks and cones are grasped from known global coordinates and placed on a foundation of half cones which guide the first layer of blocks. The construction area used in our experiments is shown in Figure \ref{fig:construction_area_wide_view}. %
Each waypoint sets a goal location for a mixed set of PID controllers that manage the vehicle's position and rotation.

\begin{figure}[h]
    \centering
    \includegraphics[width=0.95 \linewidth]{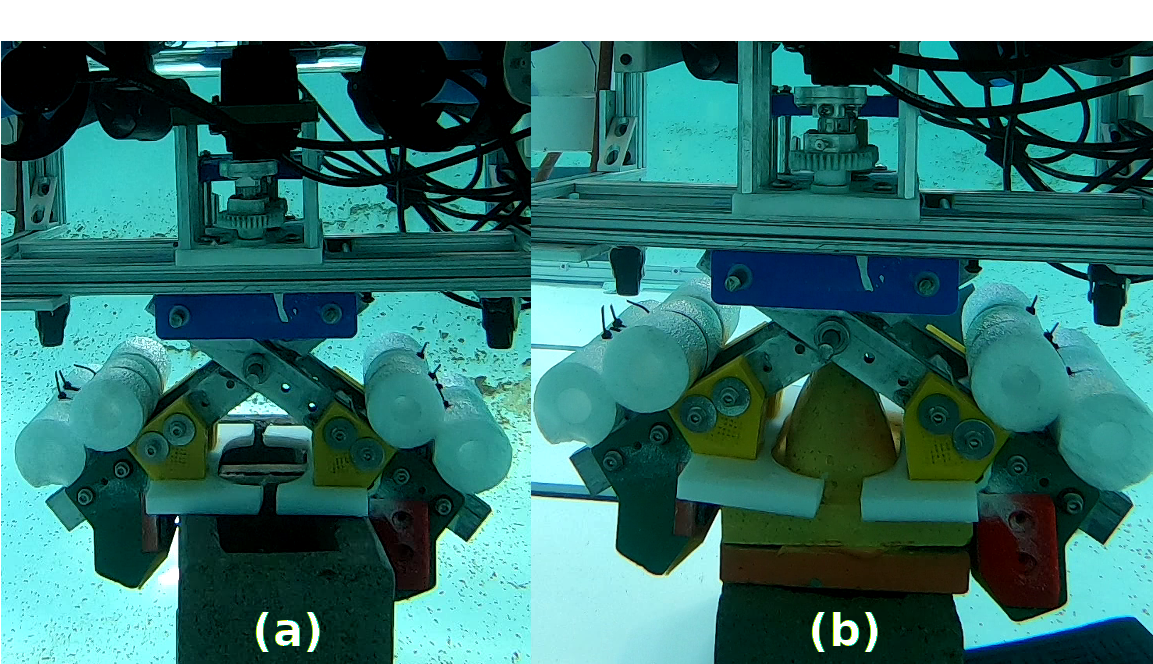}
    \caption{Plunging grasp on a block (a) and on a cone (b). The closing action of the jaws of the manipulator orients the AUV parallel with the block.}
    \label{fig:sliding-grasps}
\end{figure}

The AUV exploits its relatively small weight compared to the blocks and cones in water using a ``plunging grasp" behavior. Because the cement blocks are heavy relative to the AUV, it would require large grasp forces and complex counter-steering to force the blocks to comply with the AUV's manipulator while it grasps. Instead, we allow the AUV to comply with what it grasps. This setup allows a simple upward thrust combined with the strong closing action of the jaws to bring the AUV into a reliable alignment with the object it grasps.

After positioning above its target, the AUV disengages its PID controllers for all but its z-axis. After disengaging the controllers, the vehicle turns on a fixed upward thrust. It forces the end stop plates against the upper surface of its target while the manipulator is fully open. The tip of the cones slides into the hole in the end stop plate, adding extra error correction. The end stop plates are made of a slippery plastic material, allowing the vehicle to freely rotate on its yaw axis. As the jaws close, the vehicle is rotated to align with the block or cone. After the manipulator is fully closed, the PID controllers are re-engaged and the vehicle begins a buoyancy change behavior as described in \ref{subsec:active-ballasting}. Figure~\ref{fig:sliding-grasps} shows the plunging grasp behavior as the end stop plates slide along the top surface of both a cone and a block.

For the relatively heavy blocks, the vehicle executes a ``bailing release" maneuver in which the negative buoyancy of the robot-block system is exploited. The ballasts are fully emptied and the PID controllers of the vehicle are disengaged. The weight of the system draws the block down onto the cones below. The vehicle uses a slight upward thrust to speed the process and help prevent jamming of the block on the cones. The bailing release maneuver limits the forces placed on the structure as the block slides into place and allows the AUV to disengage without having to fight the excess positive buoyancy.

The cones are grasped with a higher degree of error correction than the blocks which allows a simpler strategy to place them onto the structure. To release the cones, the vehicle hovers above the structure and begins opening its manipulator. As the cone falls from the jaws of the manipulator, it releases the stored air in its ballast chambers. The cones are relatively light compared to the cement blocks, so the shock on the structure as they fall into place is not destabilizing. Instead, keeping the AUV distant from the structure as the cone is dropped allows the momentum of the cone to help fight jamming and prevents possible interference between the AUV and the structure.

\section{Scaling buoyancy enabled construction}

\begin{figure}
    \centering
    \includegraphics[width=\linewidth,height=1.75in]{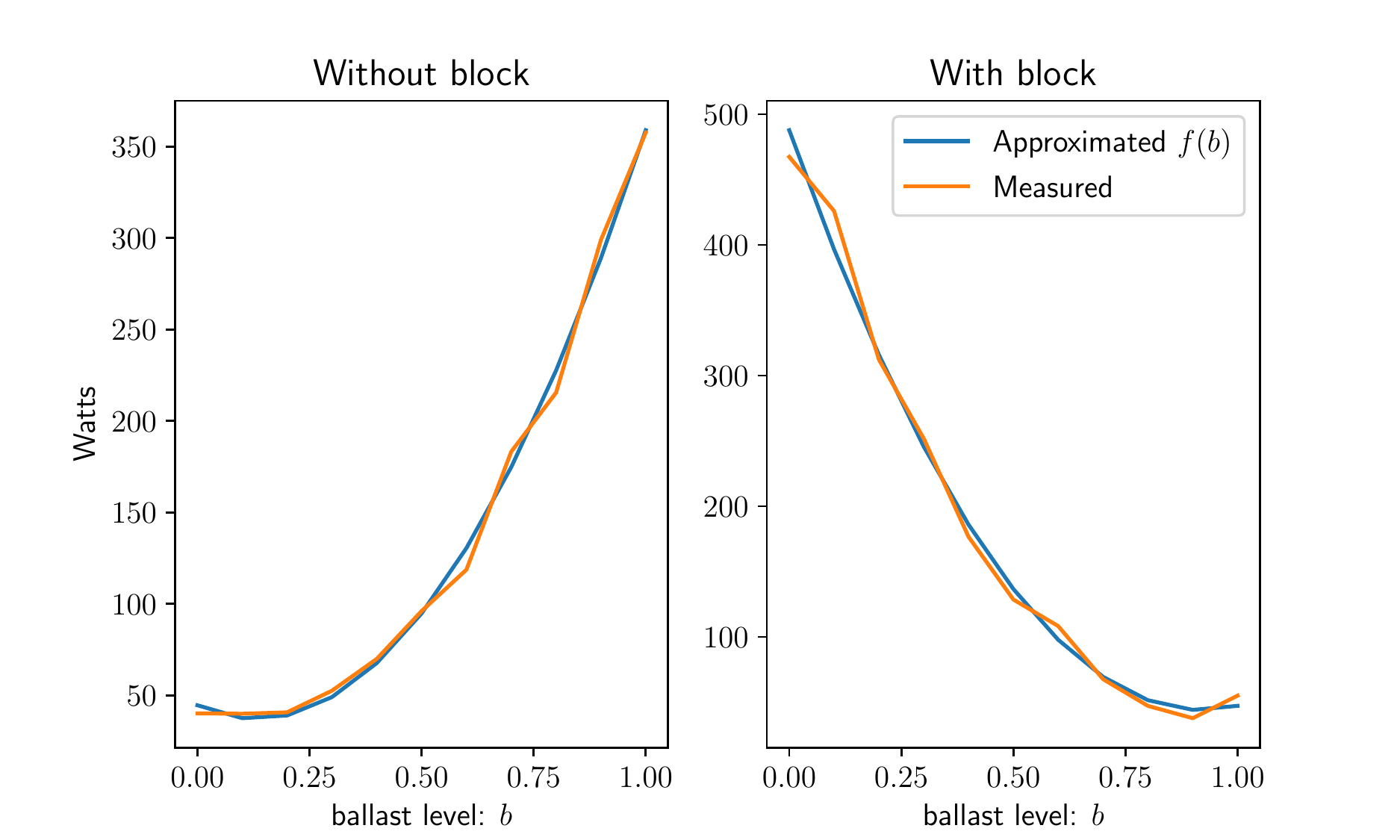}
    \caption{Energy cost of maintaining depth with increasing positive buoyancy while holding a block (a) and decreasing positive buoyancy without a block (b).}
    \label{fig:buoyancy-cost-tradeoff}
\end{figure}

Planning the construction of large scale structures using two resources presents a novel problem for construction planning. The energy efficiency of our thrusters is nonlinear in their speed. This means that the problem of allocating the amount of buoyancy, $b$, for each block is nontrivial. In this section, we explore an idealized construction problem in which buoyancy is allocated to minimize a model of the amount of battery power consumed to hold the vehicle at the desired depth. We use this model to approximate the number of times the vehicle's battery must be recharged for increasingly large structures with and without buoyancy. The convex program formulation and results in Figure~\ref{fig:buoyancy-cost-tradeoff} appeared in our previous workshop paper~\cite{Lensgraf2022StoneClawWorkshop}.

Positive buoyancy changes consume compressed air and holding the vehicle at depth using thrusters consumes battery power. A change in buoyancy at the beginning of a motion reduces the battery cost of holding depth while carrying a block. Keeping excess positive buoyancy after releasing a block increases the battery cost of navigating to the next pickup location. For construction tasks where blocks are picked up far away from where they are placed on the structure, the cost of holding the vehicle at depth while navigating to and from the structure dominates. 

If we assume that the time required to navigate between points is proportional to the distance, then the energy cost of holding the vehicle at depth over the motion is as well. In this case, a buoyancy change can be represented by a scalar that reduces the energy cost of traversing between points.

To explore the trade off between compressed air use and battery cost, we conducted an experiment in which the AUV lifted a block using varying levels of the parameter $b \in [0.0,1.0]$ during its positive buoyancy change procedure. Figure~\ref{fig:buoyancy-cost-tradeoff} shows that the cost trade-off of changing buoyancy is roughly quadratic both while holding a block and not. We use a quadratic fit of this trade off data to define a convex program for allocating buoyancy.

Let the instantaneous cost to hold depth with a block at a buoyancy level $b$ be $f_+(b)$ and $f_-(b)$ without. $f_+(b)$ and $f_-(b)$ are polynomial approximations as shown in Figure~\ref{fig:buoyancy-cost-tradeoff}. The energy cost to transport a block $d$ meters can then be approximated as $f_+(b)d\frac{1}{v}$ with average velocity $v$. 

We can idealize the construction process as traversing a set of distances $\hat d = {d_1,\dots,d_n}$ where $d_i$ is transporting a block to the structure for odd $i$ and returning to the pallet without a block otherwise. Now, let $\Delta = [\delta_1,\dots,\delta_{n}]$ be the changes in buoyancy after each action. When picking up a block at position $i$, $\delta_i$ corresponds to a positive buoyancy change, and negative otherwise.

To constrain our convex program, we can use a lower triangular matrix, $\mathbf{M}$, filled with alternating positive and negative ones to total the buoyancy level at every location. By removing the even numbered columns from $\mathbf{M}$ we get $\mathbf{M^{\prime}}$ which can be used to total the amount of compressed air used. We represent the SCUBA tank capacity by the constraint $\mathbf{M^{\prime}}\Delta \leq C$, $C \in (0, \infty)$. $0 \leq \mathbf{M}\Delta \leq 1$ ensures that the tank is never filled past $b=1$ or depleted past $b=0$.

Finally, $\mathbf{E}(\mathbf{M}\Delta)$ can be defined to predict the total energy cost: $E(\mathbf{M}\Delta) = \sum_{i=1}^n f_i((\mathbf{M}\Delta)_i)\frac{d_i}{v}$ where $f_i = f_+$ if $i$ is odd and $f_-$ otherwise. Because $\mathbf{E}$ takes the form of a linear combination of convex polynomials, it is itself a convex function.

\begin{equation}
  \label{eq:opt-problem-min}
  \begin{aligned}
  & \underset{\Delta}{\text{min}}
  & & \mathbf{E}(\mathbf{M}\Delta) \\
      & \text{subject to} & & 0 \leq \Delta \leq 1 \\
      & & & \mathbf{M^{\prime}}\Delta \leq C \\
      & & & 0 \leq \mathbf{M}\Delta \leq 1
  \end{aligned}
\end{equation}

To solve this problem, and verify the formulation, we used the disciplined convex programming solver CVXPY~\cite{cvxpy}.

Consider the case where the AUV is tasked with building the base row of a wall using blocks from a single pallet located near one side. Assuming the vehicle moves at about 0.5 meters per second when transporting a block, and that we can use one fully charged 3 liter tank which can offset about 50 blocks, how many times must we charge the vehicle's battery to build the base row starting near the pallet and moving farther away? Table~\ref{tab:buoyancy-gains} shows the result of using our convex program for increasing lengths of the row. 

\begin{table}[h!]
\centering
\caption{Approximate number of times the vehicle's \SI{230}{Wh} battery must be charged to place a row of blocks of increasing length.}
\label{tab:buoyancy-gains}
\begin{tabular}{ | c c c | }
\hline
 Blocks long & Charges with buoyancy & Charges without  \\ 
 \hline
 10 & 0 & 0 \\  
 50 & 0 & 0 \\
 250 & 4 & 16 \\
 500 & 21 & 64 \\
 \hline
\end{tabular}
\end{table}

These results show that adding the option to allocate buoyancy significantly reduces the number of times the vehicle's battery must be recharged throughout a construction process. The advantage of battery charges versus compressed air refills depends on the specific deployment scenario.

\section{Construction Trials}
\label{sec:construction-trials}

\begin{figure}[h!]
    \centering
    \includegraphics[width=\linewidth]{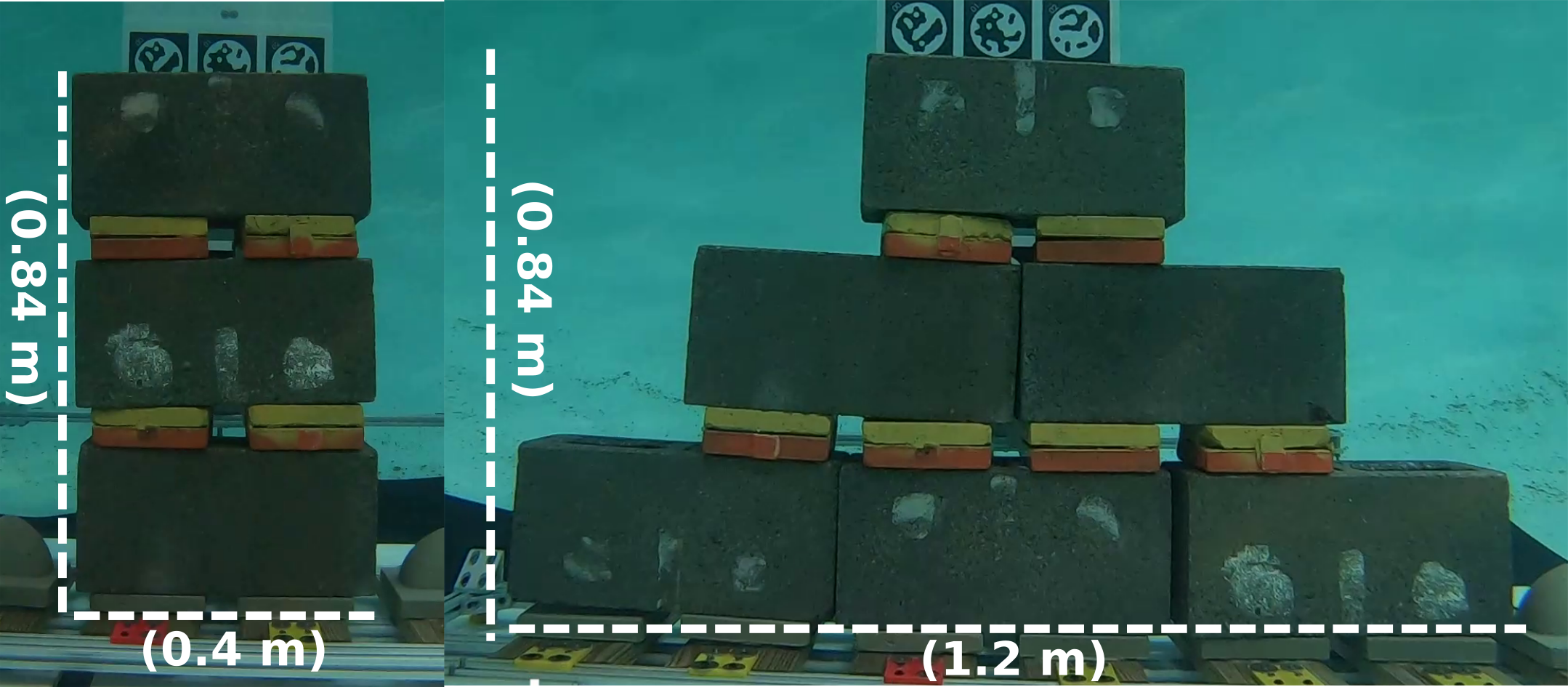}
    \caption{Two structures completed by our AUV weighing \SI{54}{kg} and \SI{100}{kg} (\SI{41}{kg} and \SI{75}{kg} in water).}
    \label{fig:structures}
\end{figure}

\begin{table}[t]
\centering
\caption{Average ratio of time and power use for executing each behavior during construction trials.}
\label{tab:power-and-time}
\begin{tabular}{ | c c c | }
\hline
 Behavior & Percent time & Percent power use  \\ 
 \hline
 Grasping block        & 5.8\%    & 9.9\%  \\
 Adding buoyancy       & 8.1\%    & 7.5\%  \\
 Placing block         & 6.2\%    & 5.2\%  \\
 Transporting block    & 44.5\%   & 47.0\% \\
 Returning to pallet   & 35.5\%   & 30.3\% \\
 \hline
\end{tabular}
\end{table}

To validate our system, we deployed it in an indoor swimming pool at about \SI{4}{m} deep. While the deployment area was controlled, periodically harsh caustics from sunlight tested our system's robustness to sub-optimal lighting. 

A fiducial marker provides global position information, and a second fiducial marker rigidly fixed to an aluminum and cement foundation provides the local position information for the construction area: the location of the slots on the foundation and the pickup locations. The foundations and block pickup locations are placed on a $2.5\times2$ meter platforms to provide a flat work surface. 

Our AUV completed three test structures: a seven component column, a nine component pyramid base and a twelve component pyramid. The column and pyramid are shown in Figure~\ref{fig:structures}. The column shows that the robot is able to place the heavy blocks without pulling down the structure. The two pyramid-like structures mimic the internal areas of a wall and show that the error correction allows the AUV to place adjacent cement blocks without jamming. 

For each manipulation, we set $b = 0.8$. The seven piece column took about 30 minutes and used \SI{51}{Wh}, the nine piece pyramid base took 38 minutes and used \SI{50}{Wh}, and the twelve piece pyramid took 45 minutes and used \SI{67}{Wh}. Manipulating each cinder block took about 60 PSI from the onboard SCUBA tank to achieve the $b = 0.8$ ballast level and the cones took about \SI{10}{PSI} of the \SI{3000}{PSI} max pressure.

We recorded the power use of the electronics system, including onboard sensing and thrusters. On average, each placement took 2 minutes 38 seconds and consumed 1.6\% of the battery.  %
Table \ref{tab:power-and-time} shows the breakdown of the average amount of time and power used during the construction of three trial structures.

Of 40 manipulations attempted during construction trials, one failed while grasping a cinder block, another failed from a missed drop of a cone and a third by missing the slots during a bailing release action. This left us with a 92.25\% success rate for component manipulation. %

\begin{figure}[h!]
    \centering
    \vspace{-1em}
    \includegraphics[width=\linewidth]{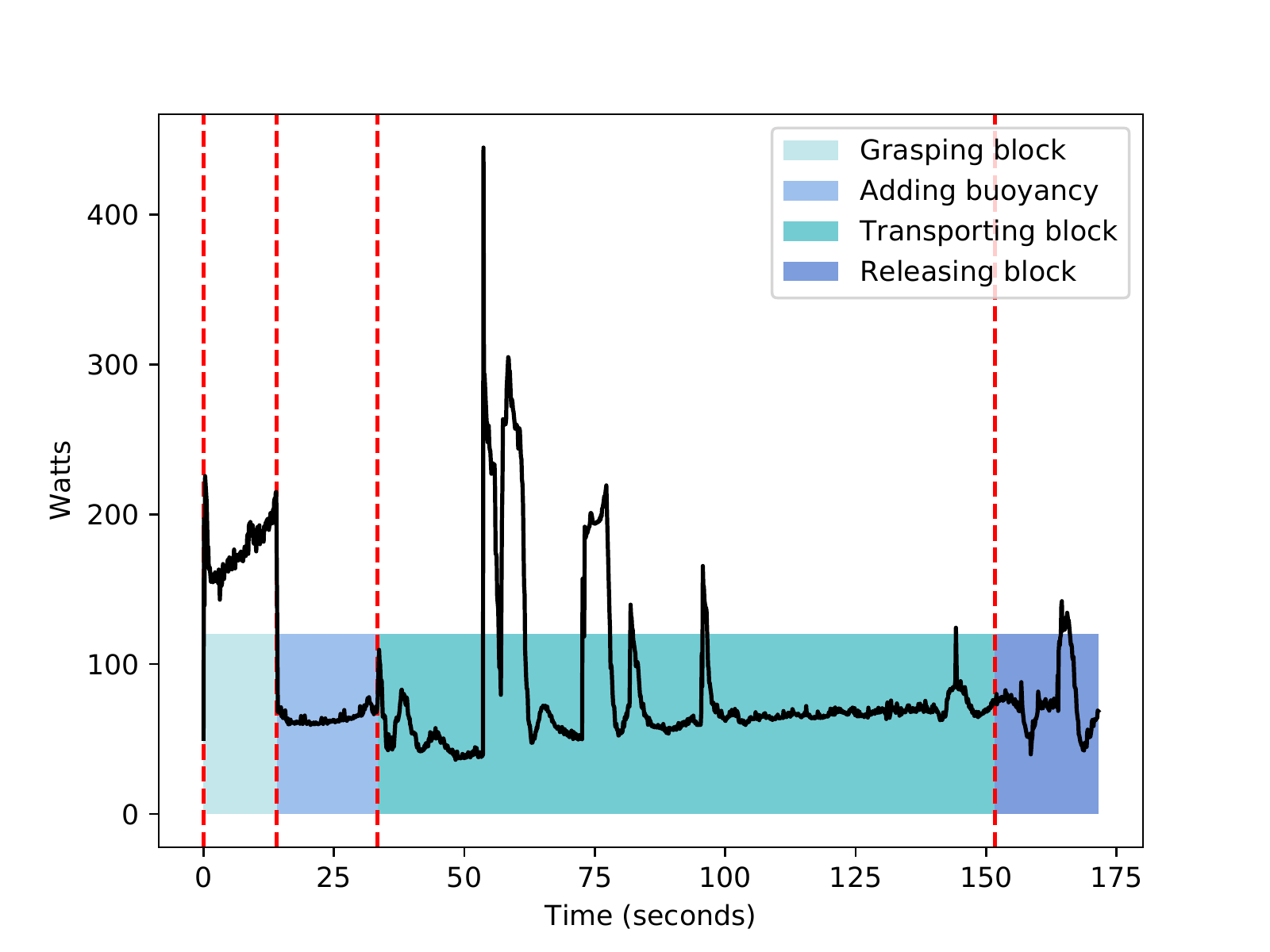}
    \vspace{-1em}
    \caption{Battery power used while transitioning through the phases to place a block.}
    \label{fig:energy_and_phase_timeline}
\end{figure}

Figure~\ref{fig:energy_and_phase_timeline} shows the behaviors the AUV proceeds through to move a block from the pickup location to the structure. The polynomial approximation of the cost to hold the AUV at depth predicts that the average cost during the ``transport" phase would be \SI{50}{W}. The real value was \SI{81}{W}. The spikes of current draw during the transit phase result from the response of the AUV's controllers as the set point is moved.

\section{Conclusions and Future Work}

This paper presents a first exploration into the construction of cement blocks structures using active ballasting. Our system is the first free-floating autonomous construction system to build with cement blocks in air or water. Our system is able to build structures of up to 12 components weighing \SI{100}{Kg} (\SI{75}{Kg} in water). To improve our system towards large scale, real world utility, we plan to address the following key challenges.

\textbf{Reliance on computer vision.} To globally localize in the platform reference frame, our AUV relies on computer vision to sense fiducial markers. While this strategy works well in controlled, clear waters, using vision to sense distant targets in real bodies of water is unreliable. In future work, we will explore hybrid visual and acoustic strategies for sensing the vehicle's position. 

\textbf{Sensing structure state.} Our current system has no direct sense of the state of the structure. This fundamentally limits the scale of structures the system can achieve. Even with a high probability of success for each manipulation action, probability of success for the structure rapidly becomes low without error recovery. In the next iteration of our system, we will explore ways to simply and reliably sense whether placement tasks are successful and recover if possible.

\textbf{More expressive building materials.} Our current building system works only in two dimensions. This is both due to the specific way we localize the AUV and the lack of right angled joining pieces. In future work, we plan to adapt both our AUV system and our building materials to allow right angles in the structures.

\textbf{Adapting design and control to external disturbances} In shallow-water marine deployments, external disturbances from waves and swells buffet the AUV. Model predictive control techniques to limit the effects of waves on an AUV can help alleviate position uncertainty, but recent experimental work by \textcite{Walker2021stationkeeping} shows up to an \SI{0.6}{m} root mean squared error when attempting to remain in a fixed position. This error is larger than the \SI{2.5}{cm} error correction capacity of our building components. To harden our system to real-world external disturbances, we plan to increase the acceptance area of our construction components and employ rapid release strategies to quickly place components opportunistically as the AUV's position oscillates.

\printbibliography

\end{document}